# Benchmarking Single-Pose Docking, Consensus Rescoring, and Supervised ML on the LIT-PCBA Library: A Critical Evaluation of DiffDock, AutoDock-GPU, GNINA, and DiffDock-NMDN

**Youssef Abo-Dahab[1], Xiaoiang Xiang[1,2], Joanne Chun[1,2], Liang Zhao[1,2]**

[1] Department of Bioengineering and Therapeutic Sciences, Schools of Pharmacy and Medicine, University of California, San Francisco, San Francisco, CA, USA
[2] UCSF–Stanford Center of Excellence in Regulatory Science and Innovation (CERSI), San Francisco, CA, USA

## Abstract

**Background:** Virtual screening (VS) prioritizes compounds for testing but its impact hinges on the method used. Recent Artificial Intelligence (AI) based components (e.g., DiffDock for pose generation; NMDN-DiffDock and GNINA for rescoring) report strong benchmark numbers, yet their practical value on experimentally curated libraries remains uncertain.

**Objective:** Evaluate virtual screening on LIT-PCBA library across 15 targets using two methods for docking/pose generation including AutoDock and DiffDock, then rescored with GNINA and NMDN (DiffDock-NMDN); test whether calibrated rank-based consensus with score filtering improves early enrichment over any single scorer.

**Methods:** We docked 578,295 ligand–target pairs (10,008 active; 568,287 inactive). In the AutoDock pathway, each ligand had 10 runs and the best-affinity pose was kept; in the DiffDock pathway, 20 poses were sampled and the highest-confidence pose was retained. Poses were rescored with GNINA and NMDN. We produced rankings from individual scorers, multiple consensus schemes for each pathway, and a Global Consensus that merges both pathways. Primary metrics were EF1% and EF10%, with ROC-AUC and BEDROC(α=20). We also trained multiple Machine Learning (ML) models using various architectures.

**Results:** GNINA rescoring of AutoDock poses (AutoDock-GNINA) was the most dependable single method (median EF1% = 2.14, Precision =1.85%, Recall = 2.02%, balanced accuracy=50.5%), outperforming DiffDock-GNINA (median EF1% = 0.84) and both baselines (AutoDock median EF1% = 1.10; DiffDock median EF1% = 1.17). consensus ranking improved robustness: yielded median EF1% = 1.8 in both pathways and rescued 2 targets that AutoDock-GNINA alone missed; Global Consensus reached the highest median EF1% among consensus variants (1.9), but still less than AutoDock-GNINA alone. Notably, AutoDock-GNINA uniquely rescued OPRK1 (EF1% = 12.5) where all DiffDock-based variants failed (EF1% = 0). Throughput favored AutoDock over DiffDock for speed and cost, 4 to 8 times cheaper. Supervised ML re-ranking delivered the largest gains: our best ML model achieved EF1% = 4.49 (+110% improvement over AutoDock-GNINA's 2.14) and improved balanced accuracy to 55.4% from 50.5%.

**Conclusion**: Even though AutoDock-GNINA offers the best balance between accuracy and cost in our experiments, it was only slightly better than random screening. DiffDock's single-pose predictions proved less reliable for downstream rescoring, while NMDN delivered inconsistent, target-specific gains. When carefully implemented, consensus docking improved enrichment robustness, particularly within the DiffDock pathway. Supervised ML models can significantly boost docking enrichment, if experimental data is sufficient. Overall, no single docking technique works on all targets. Therefore, we believe that employing a rigorously tested docking technique, whose limitations are well understood, is preferable to relying on an unpredictable novel method that may yield inconsistent results.

## Introduction

Virtual screening (VS) has become an indispensable component of modern, cost-effective drug discovery pipelines.[1] The central premise of VS is to computationally evaluate huge libraries of chemical compounds—often millions or more—to identify a small, enriched subset of molecules with a high likelihood of binding a biological target of interest.[2] By pre-filtering chemical spaces in silico, VS dramatically reduces the number of compounds that must be tested in expensive and time-consuming experimental high-throughput screening (HTS).[3] The predictive power of VS methods, however, depends critically on the quality of the algorithms and the rigor of the datasets used to evaluate and validate their performance.

Historically, the field relied on artificially constructed benchmarking sets such as the Directory of Useful Decoys (DUD), the Maximum Unbiased Validation (MUV) dataset, and its successor, DUD-E. [4] However, numerous investigations have demonstrated that these benchmarks are unfortunately compromised by both obvious and hidden chemical biases. [5,6] These flaws include analog bias, where active compounds are structurally much more similar to each other than to the inactive "decoy" compounds, and the presence of decoy artifacts, where models learn to distinguish actives from inactives based on simple physicochemical properties rather than true binding interactions. Consequently, these datasets often produce artificially inflated performance metrics, overestimating the true accuracy of VS methods and providing a misleading picture of methodological progress.

In this study, we sought to evaluate docking and scoring methods under more realistic conditions by using a dataset that closely approximates high-throughput screening conditions, specifically the LIT-PCBA (Literature-derived PubChem Bioassay). It was constructed using data from 149 dose-response PubChem bioassays, incorporating thousands of experimentally confirmed active compounds and hundreds of thousands of experimentally confirmed *inactive* compounds, rather than computationally generated decoys. This approach was designed to mimic the conditions of a real-world experimental screening deck, characterized by very low hit rates. [5,6] By testing methods on LIT-PCBA, we aimed to obtain a more stringent and practical assessment of their enrichment capabilities.

We evaluated a representative set of docking workflows spanning both classical and recent AI-driven approaches. We compared AutoDock-GPU (a modern GPU-accelerated implementation of AutoDock 4.2, referred to simply as “AutoDock” hereafter) and DiffDock (a diffusion generative model for blind docking) for pose generation. For scoring and ranking the poses, we examined the performance of AutoDock’s own physics-based scoring, DiffDock’s internal confidence score, GNINA (a 3D convolutional neural network scoring function), and NMDN (a Normalized Mixture Density Network rescoring approach). This framework allows us to test two key questions: (1) Do state-of-the-art machine learning-based docking/scoring methods outperform a traditional physics-based docking tool on an unbiased experimental benchmark? (2) Does combining different scoring methods in a rank-based consensus improve early enrichment of actives compared to the best individual method? Additionally, we explored a supervised learning approach by training machine-learning models on the docking results to determine if a learned re-ranker can further boost enrichment.

### Early Enrichment Metrics: EF1% and EF10%

Enrichment Factor (EF) is a standard metric used to assess the performance of a virtual screening method by measuring its ability to preferentially rank active compounds at the top of a sorted list. The EF at a specific percentage of the dataset denoted as $EF_{X\%}$, is defined as the ratio of the observed enrichment to the expected random enrichment.

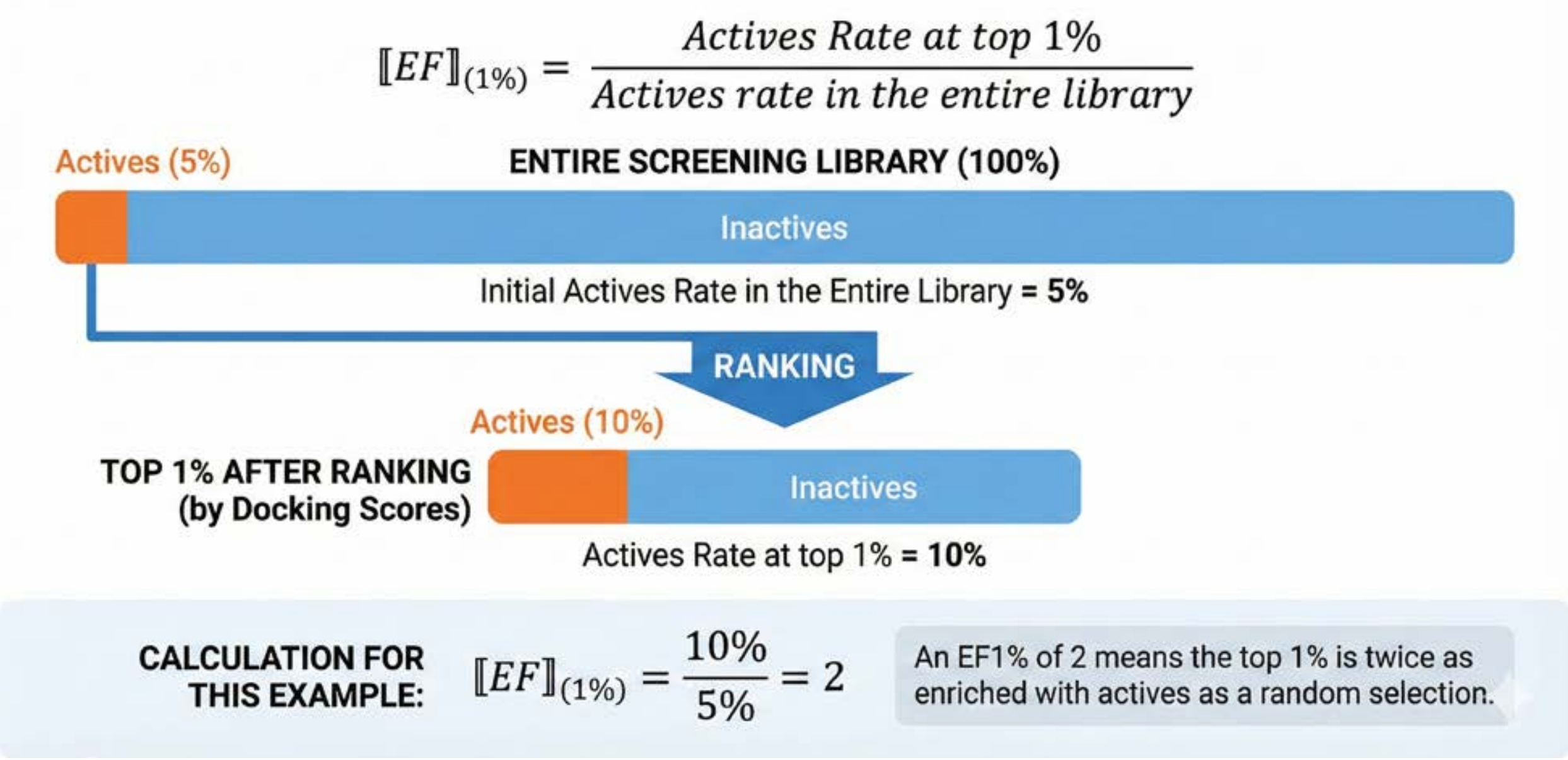

**Figure 1.** illustrates an example of EF1% calculation, where the percentage of actives in the entire library is 5%. However, after docking and ranking the top1% have 10% actives, which is twice the library rate, that's why we say it has an EF1% of 2.

A higher EF value indicates a more effective method for prioritizing active molecules. EF1% is particularly useful for evaluating a model's ability to find highly potent compounds that would be prioritized for immediate experimental validation, whereas EF10% provides a broader view of a model's overall early enrichment. [7].

## AutoDock

In this work, we employed AutoDock-GPU, a modern, hardware-accelerated implementation of the classic AutoDock 4.2 algorithm. The original AutoDock, based on the Lamarckian Genetic Algorithm (LGA) for conformational searching, is computationally intensive and inefficient for large-scale virtual screening on standard CPUs. AutoDock-GPU addresses this limitation by leveraging Graphics Processing Units (GPUs) through OpenCL and CUDA, enabling the simultaneous processing of thousands of ligands–receptor poses. According to Santos-Martins et al. (2021), AutoDock-GPU delivers runtime speedups of approximately 30×–350× with the Solis–Wets local search and 2×–80× with the ADADELTA local search, depending on ligand size and search parameters. Additional studies have explored the computational performance and workflow optimization of AutoDock-based pipelines in high-throughput screening settings, highlighting the importance of efficient parallelization and scalable docking strategies. [8,9] For simplicity, AutoDock-GPU is referred to as AutoDock throughout this manuscript. [10]

**LIT-PCBA Performance (by Proxy):** Across the 15 targets of the LIT-PCBA dataset, AutoDock Vina achieves a median EF1% of **1.3** and a median AUROC of **0.61**.[1]

## DiffDock: Using Machine Learning for Docking

Unlike traditional physics-based docking that AutoDock uses, DiffDock frames molecular docking as a generative modeling problem, using a diffusion process over the non-Euclidean manifold of ligand poses—including translational, rotational, and torsional degrees of freedom. The model is trained to reverse this diffusion, starting from random (noisy) poses and iteratively denoising them to produce likely binding conformations. This enables blind docking—predicting binding poses without prior knowledge of the pocket. In benchmarks, DiffDock achieved a 38 % top-1 success rate (RMSD < 2 Å) on PDBBind, significantly outperforming traditional docking (~23 %) and regression-based deep learning (~20 %) methods. [11]

Despite this impressive reported performance, a recent critique suggests DiffDock's results may be overly optimistic due to training set biases. [12] The analysis found that many PDBBind test cases had nearly identical protein–ligand examples in DiffDock's training set ("near neighbors"), and DiffDock performed much better on those cases. When truly novel complexes (no close analogs in training) were evaluated, success rates dropped by ~40 percentage points. This implies that DiffDock may partially memorize training examples rather than learn generalizable docking principles [12]. Thus, it remained unclear how well DiffDock would work on novel targets or in VS scenarios with no homologous complexes – one motivation for testing it on LIT-PCBA.

## GNINA: Using Machine Learning for Rescoring

GNINA is a docking scoring function that augments the AutoDock Vina scoring approach with 3D convolutional neural networks (CNNs) for pose scoring and ranking. GNINA places the protein–ligand complex in a 3D grid (voxel space) and uses CNN filters to learn spatial patterns indicative of binding, rather than relying purely on empirical energy terms. This data-driven approach can capture subtle stereochemical and recognition features that might be missed by simpler scoring functions. [1]

**Performance on LIT-PCBA according to Sunseri, J., & Koes, D. R. (2021):** Across the combined DUD-E and LIT-PCBA benchmarks, the default GNINA scoring function was found to outperform Vina on 89 of 117 targets. On the LIT-PCBA benchmark specifically, it achieved a median EF1% between 1.88 and 2.58 (depending on the mode), which is roughly twice the early enrichment performance of Vina (0.90). The models achieved a median ROC-AUC of approximately 0.61–0.62 on LIT-PCBA. [1]

## DiffDock-NMDN Protocol

Given DiffDock's strength in pose generation, researchers have explored augmenting it with alternative scoring. Xia et al. (2025) introduced a two-stage process operates as follows:

1. Pose Generation: For each protein-ligand pair, DiffDock is used to sample protocol called DiffDock-NMDN to improve DiffDock's screening enrichment [13]. In the first stage, DiffDock generates multiple potential binding candidate poses across the entire protein surface per ligand in a blind docking fashion.

2. Rescoring with NMDN: The generated poses are then evaluated and ranked in the second stage, using Normalized Mixture Density Network (NMDN) score. NMDN is a distinct deep learning-based scoring function that learns the probability density distribution of distances between protein residues and ligand atoms to estimate binding strength, producing a "pKd-like" score for binding. Using this DiffDock+NMDN pathway, Xia et al. reported an average EF1% of 4.96 on LIT-PCBA – among the highest seen on that benchmark [13]. In our study, we adopt the NMDN rescoring concept for both DiffDock and AutoDock pathways. However, unlike Xia et al., who sampled many DiffDock poses and used NMDN to select the best, we used a single-pose evaluation: each ligand contributes only one pose (the top-scoring pose from AutoDock or the highest-confidence pose from DiffDock), which we then rescore with NMDN. This allows a fair comparison of pathways under consistent single-pose conditions.

**Table 1**. Summary of methods used for docking and scoring in this study

| Method | Used for docking and pose generation | Used for scoring and ranking |
|---|---|---|
| AutoDock | Yes | Yes |
| DiffDock | Yes | Yes |
| GNINA | No | Yes |
| NMDN | No | Yes |

AutoDock and DiffDock perform both docking (pose generation) and scoring using their internal scoring functions. GNINA and NMDN are applied only as rescoring methods on poses generated by the docking programs.

## Methodology:

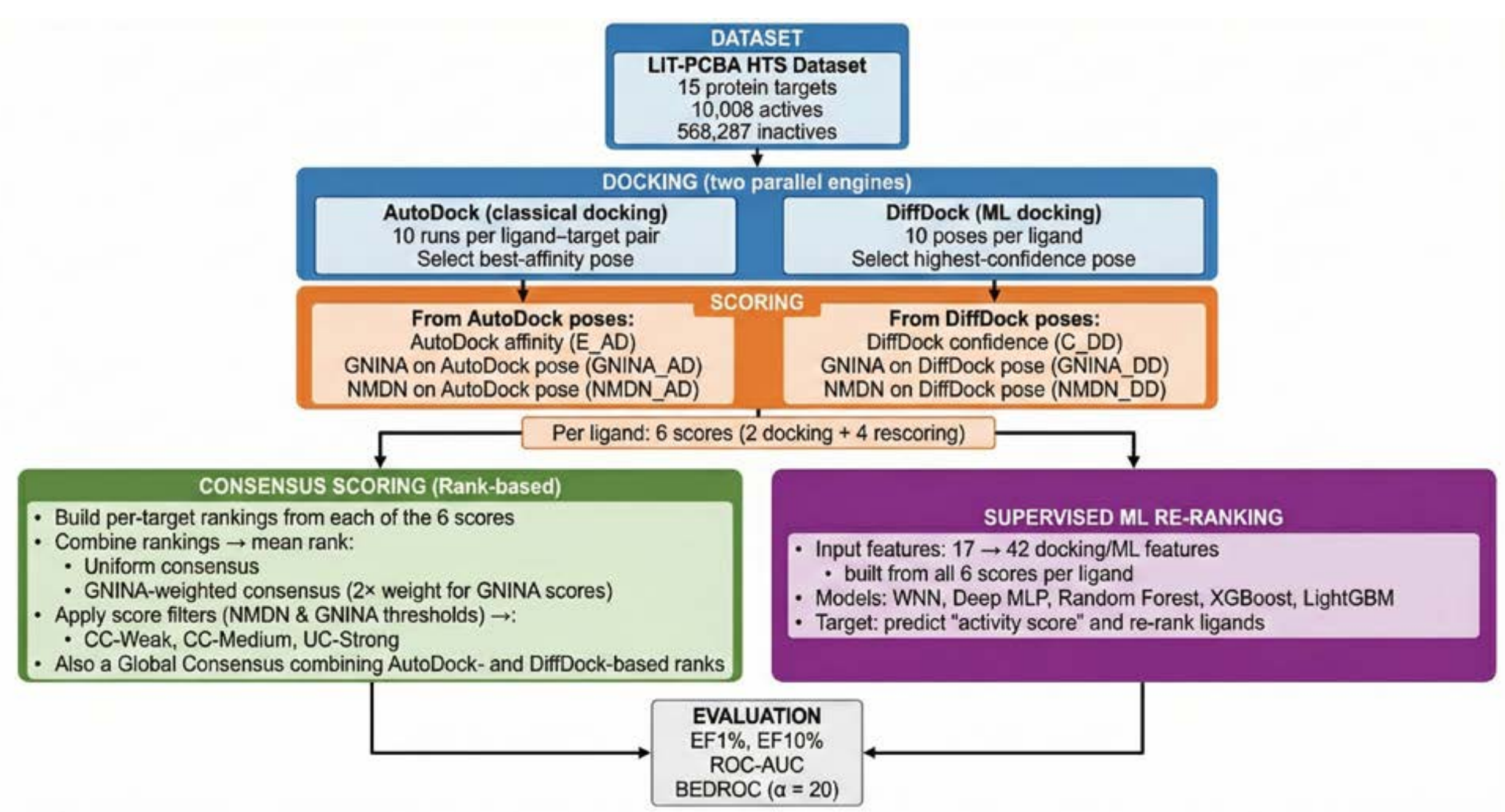

**Figure 2. Methodology flowchart of the virtual screening pipeline.** This summarizes the virtual screening pipeline used in the study. The LIT-PCBA dataset (15 targets; ~578k ligands) is screened using two parallel docking engines: classical AutoDock and ML-based DiffDock. For each ligand, a single representative pose is selected (best-affinity pose for AutoDock; highest-confidence pose for DiffDock) and evaluated with multiple scoring functions, including the native docking scores and ML re-scorers (GNINA and NMDN), yielding six scores per ligand. Ligands are then prioritized using rank-based consensus scoring with optional score filtering, as well as supervised machine-learning re-ranking models trained on docking and rescoring features. Performance is assessed using early-enrichment and ranking metrics (EF1%, EF10%, ROC-AUC, and BEDROC), enabling comparison of individual methods, consensus strategies, and ML re-ranking approaches. Detailed methodology explanation is further discussed.

## Datasets and Targets

We evaluated all methods on 15 diverse protein targets drawn from the LIT-PCBA benchmark collection. In total, **578,295** ligand–protein pairs were docked (10,008 known actives and 568,287 known inactives across the 15 targets). The number of molecules per target ranged from **4,245** (for the smallest target, TP53) up to **197,274** (for the largest, OPRK1). For seven of the targets (KAT2A, IDH1, GBA, FEN1, ADRB2, VDR, and PKM2), we down sampled the inactive set to **5%** of its full size (while keeping all actives) due to computational constraints. All other targets used the complete set of inactives. Without this subsampling, processing the full library would have entailed docking ~2.5 million inactive compounds, which was infeasible within our computational budget.

### Molecular Docking and Scoring Pathways

We developed two distinct computational pathways to identify potential binders from a library of molecules. The first pathway, AutoDock, was employed for a traditional molecular docking approach. Each ligand underwent 10 independent docking runs, and the pose with the most favorable binding affinity was selected for subsequent analysis. The resulting poses were then re-scored using two rescoring methods: GNINA, which provides a CNNaffinity score (higher values indicating better binding), and NMDN, which predicts a pKd value (higher values indicating stronger affinity).

The second pathway utilized the DiffDock for docking. For each ligand, 20 poses were generated, and the pose with the highest confidence score was selected. Similar to the AutoDock pathway, these selected poses were then re-scored using GNINA's CNNaffinity and NMDN's predicted pKd.

### Ranking and Consensus Scoring

For each target and each scoring function, we ranked all ligands by their score (actives and inactives intermingled) to evaluate enrichment. Lower AutoDock binding energies indicate better rank (more negative = tighter binding), whereas for DiffDock, GNINA, and NMDN, higher scores indicate better rank. To facilitate comparisons, we converted all scores into ranks (1 = best) for each scoring method within each pathway.

We then explored consensus scoring strategies that combine multiple ranking lists to produce a single, potentially more robust ranking. We considered a simple averaging of ranks across methods (which treats each scoring method equally) and a weighted averaging that gives more emphasis to the ML-based scorers (GNINA and NMDN). Specifically, we evaluated:

- Unweighted averaging: All scoring methods were given equal weight in the consensus contributed equally. For a given ligand, we took its rank from AutoDock (or DiffDock) score, its rank from GNINA, and its rank from NMDN, and computed the simple average of these ranks.

- Weighted Consensus (Also referred to in the is paper as Calibrated Consensus, or CC for short): GNINA and NMDN were assigned a double weight (2×) compared to the base docking method (AutoDock or DiffDock, weighted 1×).

$$average\ rank = \frac{W_{NMDN} Rank_{NMDN,i} + W_{GNINA} Rank_{GNINA,i} + W_{Baseline_docking} Rank_{Baseline_docking}}{W_{NMDN} + W_{GNINA} + W_{Baseline_docking}}$$

Equation 1: Illustrates how to calculate a weighted average rank by changing the W of each score, in unweighted averaging W of each score will be 1.

Thus, if a molecule has a rank of 15 in GNINA, 76 Baseline and 939 in NMDN, the mean rank for uncalibrated method is 343.3 while for calibrated method ($W_{GNINA} = 2$) is 261.25.

### Score-Based Filtering and Enrichment Analysis

To further optimize the identification of active compounds, we investigated the effect of applying score-based cutoffs prior to ranking. Molecules with scores below a specified threshold were filtered out. Specifically, we applied several cutoff values for both NMDN_Score (900, -800, and -4000) and GNINA CNNaffinity (0.6, 0.1, and 0.0). We explored with hundreds of combinations, and we found that these cutoffs are more consistent than others. Both pathways are illustrated in Figure 3.

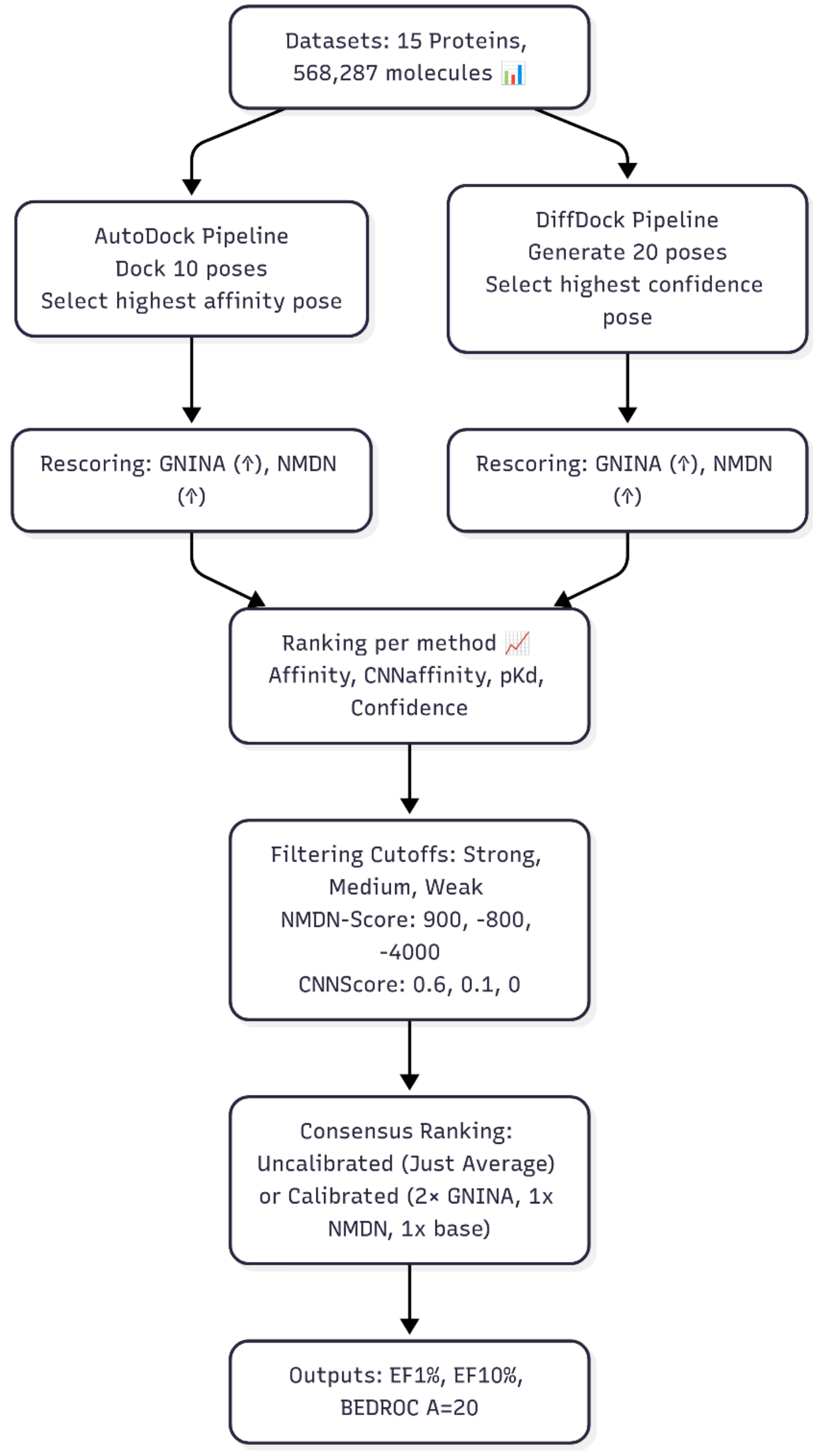

**Figure 3.** The flowchart of consensus docking.

We created 3 distinct filtering methods:

**Calibrated Consensus with Medium Filtering (CC-Medium)**: This strategy employed a medium filtering approach, retaining only molecules with an NMDN score of at least -800 and a CNNScore of at least 0.1. The consensus ranking was then generated using a calibrated weighting scheme, in which the GNINA ranking was given double the weight of the other scoring methods.

**Uncalibrated Consensus with Strong Filtering (UC-Strong)**: A more stringent filtering criterion was applied in this case, keeping only molecules with an NMDN score of at least 900 and a CNNScore of at least 0.6. This ranking was constructed using an uncalibrated consensus, where all methods were given equal weight.

**Calibrated Consensus with Weak Filtering (CC-Weak)**: This approach used the most permissive filter, with a NMDN score cutoff of -4000 and a CNNScore cutoff of 0.0. This weak filtering strategy was intended to retain a high percentage of the initial dataset, including most known active molecules. The final ranking was created using the calibrated consensus method (2x weight for GNINA).

**Global Consensus Ranking Method**: In this method, we used the same settings of the **CC-Medium** filter. (-800 NMDN-Score, and 0.1 CNNScore), and same calibration 2x GNINA 1x NMDN 1x baseline docking methods. We used these settings because they produced the best results in both pathways. So, we wanted to see if it could even improve results further.

## Machine-Learning Re-Ranking Models

In addition to the above rank-based consensus methods, we investigated whether a supervised machine learning model could learn to better distinguish actives from inactives using the various docking scores as features. We constructed several classification models trained to predict active vs inactive, effectively re-ranking the library with a learned model. Below we outline the training procedure and model details:

**Feature Engineering:** For each ligand–target pair, we computed a set of descriptive features from the docking results. We started with 17 primary features, which included the raw docking scores from AutoDock (binding energy), DiffDock (confidence score), GNINA (CNN affinity), and NMDN (predicted pKd), as well as related metrics (for example, if multiple poses or scores were considered, we could include min/mean values – though in our single-pose setup, each is just one value). We then expanded this feature set to 42 derived features by including non-linear transforms (log and square of top features), interaction terms (products or differences between scor es), and simple statistics (mean and standard deviation across methods for targets where multiple methods apply). Prior to modeling, features were scaled using a RobustScaler (which centers and scales data based on median and interquartile range) to mitigate the effect of outliers while preserving relative rankings.

**Dataset Splits:** To train and evaluate the models, we split the data by target to avoid any leakage of ligand-specific patterns across training and test. Approximately 75% of the ligand–target pairs (for each target) were used for training and 25% held out for validation. This yielded roughly ~417,000 training samples and ~139,000 validation samples overall, ensuring that no target's actives or inactives were exclusively in one set (each target's data was internally split).

**Neural Network Architectures**: We benchmarked two fully connected feed-forward designs: (i) a Wide Neural Network (WNN), front-loaded with 512 neurons in the first layer followed by 256, 128, and 1 output unit, with batch normalization and adaptive dropout (0.3→0.21→0.15) applied at successive layers; and (ii) a Deep MLP, with a narrower progression (256→128→64→1) and fixed dropout (0.3→0.2→0.1). Both networks used the Adam optimizer (learning rate = 0.001, weight decay = $1\times10^{-5}$) with a dynamic batch size (N/4, where N is training set size) and a maximum of 30 epochs. Early stopping was applied based on validation EF1%.

**Tree-Based Models:** Gradient boosting methods were evaluated using XGBoost in multiple parameterizations. The best configuration (200 estimators, depth 6, learning rate 0.05) achieved EF1% = 3.80, while shallower or faster variants delivered EF1% in the range 3.55–3.70. LightGBM was run in lambda rank mode (LambdaMART implementation). Random Forest classifiers with 100 trees and maximum depth 8 reached EF1% = 4.10, rivaling neural networks and outperforming all boosting variants.

**BEDROC and ROC-AUC**

While EF is a useful metric, it has limitations, particularly its sensitivity to the size of the evaluated fraction (x%). To provide a more robust assessment of model performance, we also utilized BEDROC and ROC-AUC.

**ROC-AUC** stands for Receiver Operating Characteristic - Area Under the Curve. It is a powerful statistical tool used to evaluate the performance of a binary classification model by measuring how well a model's scores can rank positive cases (active molecules) higher than negative cases (inactive molecules). The foundation of the metric is the ROC curve, which plots the True Positive Rate (TPR) against the False Positive Rate (FPR) at every possible score threshold. By doing this, the curve visually represents the trade-off between finding true positives and avoiding false positives. A perfect model would have a curve that shoots straight up to 1.0 TPR at 0.0 FPR, giving a perfect score of 1, while a random-guessing model follows the diagonal line, giving a score of 0.5. Any value greater than 0.5 indicates performance better than random chance. [14]
For each method we passed the ranking score function as way to calculate ROC-AUC. For example, in GNINA, the CNNAffinity score is the one used to rank molecules, so we used it to calculate ROC-AUC.

**BEDROC (Boltzmann-Enhanced Discrimination of ROC)** is a robust metric designed to assess the performance of virtual screening models by specifically measuring early enrichment. Unlike traditional metrics like ROC-AUC, which evaluate performance across the entire dataset, BEDROC places a strong emphasis on the ability of a model to rank true active compounds at the very top of a sorted list. This is particularly relevant in drug discovery, where experimental validation is resource-intensive and focuses on a small fraction of top-ranked candidates.

The BEDROC score is derived from the Robust Initial Enhancement (RIE), a metric that exponentially weights the ranks of active compounds, giving much higher scores to actives found near the beginning of the list. The final BEDROC score is a normalized version of RIE, scaled to range from 0 (random performance) to 1 (perfect performance). [7]

$$\mathrm{BEDROC} = \frac{\mathrm{RIE} - \mathrm{RIE}_{\mathrm{min}}}{\mathrm{RIE}_{\mathrm{max}} - \mathrm{RIE}_{\mathrm{min}}}$$

$$\text{where } \mathrm{RIE} = \frac{\frac{1}{n}\sum_{i=1}^{n} e^{-\alpha x_i}}{\frac{1}{N}\left(\frac{1-e^{-\alpha}}{e^{\alpha/N}-1}\right)}, \mathrm{RIE}_{\mathrm{min}} = \frac{1-e^{\alpha R_{\mathrm{a}}}}{R_{\mathrm{a}}(1-e^{\alpha})}, \text{ and } \mathrm{RIE}_{\mathrm{max}} = \frac{1-e^{-\alpha R_{\mathrm{a}}}}{R_{\mathrm{a}}(1-e^{-\alpha})}$$

Equation 2

- RIE is the exponentially-weighted sum of the ranks of all active compounds.
- $RIE_{min}$ represents the expected RIE value for a random ranking.
- $RIE_{max}$ represents the RIE value for a perfect ranking where all active compounds are at the top.

The core of the BEDROC metric is the α parameter, which controls the magnitude of the early enrichment emphasis. A higher α value (e.g., α=20) heavily rewards models that place actives in the top 1% of the ranked list, while a lower α value makes the metric more similar to ROC-AUC. This flexibility allows BEDROC to be tailored to the specific needs of a screening campaign. BEDROC offers a more nuanced and reliable single-number metric for early enrichment compared to the cutoff-dependent Enrichment Factor (EF)

## Results

After docking the 578,295 molecules using both pathways, rescoring them with different scoring methods, evaluating EF1%, EF10%, ROC-AUC, BEDROC alpha = 20, and percentage of remaining actives (for consensus filtered methods) for each protein, the overall median and average results per method and pathway are illustrated in tables 2 and 3.

**Table 2**. Median virtual screening performance across 15 protein targets

| Pathway | Scoring Method | Median EF1% | Median EF10% | Median ROC-AUC | Median BEDROC (A=20) | Median Actives Remaining | Success times |
|---|---|---|---|---|---|---|---|
| AutoDock | AutoDock | 1.1 | 0.77 | 0.4568 | 0.06 | - | 5 |
| AutoDock | DiffDock-NMDN | 0.37 | 1.2 | 0.571 | 0.066 | - | 4 |
| AutoDock | GNINA | **2.14** | **1.63** | **0.62** | **0.12** | - | 8 |
| AutoDock | CC-Medium | **1.84** | 1.53 | - | 0.0723 | 83.3% | **9** |
| AutoDock | UC-Strong | 0 | 0 | - | 0 | 0 | 6 |
| AutoDock | CC-Weak | 1.62 | 1.53 | - | 0.0887 | **100%** | **9** |
| DiffDock | DiffDock | 1.17 | 0.83 | 0.53 | 0.0762 | - | 5 |
| DiffDock | DiffDock - NMDN | 0.673 | 1.13 | 0.5744 | 0.0642 | - | 6 |
| DiffDock | GNINA | 0.84 | 1.31 | 0.5535 | 0.0666 | - | 4 |

| DiffDock | CC-Medium | 1.8 | 1.54 | - | 0.0683 | 82% | 8 |
|---|---|---|---|---|---|---|---|
| DiffDock | UC-Strong | 0 | 0 | - | 0 | 0 | 5 |
| DiffDock | CC-Weak | 1.8 | 1.47 | - | 0.0891 | **100%** | **9** |
| GLOBAL | CC-Medium | 1.9 | 1.58 | - | 0.0634 | **69%** | 8 |

Values represent median performance across all targets. Key metrics include Enrichment Factor in the top 1% (EF1%), BEDROC, and the number of successful screens (Success times). Notably, the AutoDock pathway paired with GNINA rescoring and the CC-Medium method demonstrates the strongest and most consistent early enrichment

**Table 3**. Average virtual screening performance metrics across 15 protein targets.

| Pathway | Scoring Method | Average EF1% | Average EF10% | Average ROC-AUC | Average BEDROC (A=20) | Average Actives Remaining | Success times |
|---|---|---|---|---|---|---|---|
| AutoDock | AutoDock | 1.4 | 1.06 | 0.456 | 0.07 | - | 5 |
| AutoDock | AutoDock-NMDN | 0.556 | 1.33 | 0.570 | 0.067 | - | 4 |
| AutoDock | GNINA | **4.07** | **2.02** | **0.618** | **0.13** | - | 8 |
| AutoDock | CC-Medium | 2.62 | 1.65 | - | 0.08 | 68.4% | **9** |
| AutoDock | UC-Strong | 2.87 | 2.14 | - | 0.0037 | 4% | 6 |
| AutoDock | CC-Weak | 2.1 | 1.68 | - | 0.1 | 98.8% | **9** |
| DiffDock | DiffDock | 1.41 | 1.116 | 0.538867 | 0.086 | - | 5 |
| DiffDock | DiffDock - NMDN | 0.834 | 1.07 | 0.563873 | 0.067 | - | 6 |
| DiffDock | GNINA | 1.89 | 1.586 | 0.58562 | 0.096 | - | 4 |
| DiffDock | CC-Medium | 3.04 | 1.6 | - | 0.080 | 71.7% | 8 |
| DiffDock | UC-Strong | **3.3** | **2.05** | - | 0.010 | 4% | 5 |
| DiffDock | CC-Weak | 2.9 | 1.7 | - | 0.1012267 | **99.3%** | **9** |

| GLOBAL | CC-Medium | **3.45** | 1.63 | **-** | 0.084 | **69%** | 8 |
|---|---|---|---|---|---|---|---|

Unlike the median, the average is sensitive to outliers and can reveal exceptionally high performance on a subset of targets. This is evident in the higher average EF1% for the UC-Strong method, which performed poorly on a median basis but achieved remarkable enrichment on a few specific targets. Here again, AutoDock-GNINA achieved highest score with 4.07 EF1%

**Performance of single methods:**

**Baseline DiffDock:** DiffDock baseline predictions showed high variability across the 15 LIT-PCBA targets. DiffDock failed to retrieve any actives in the top 1% for four targets, namely ADRB2, IDH1, OPRK1, and PPARG, resulting in a relatively low success rate (33% only). Median EF1% was 1.17, and Median BEDROC was 0.076.

**Baseline AutoDock:** AutoDock baseline docking performance was worse than baseline DiffDock. Median EF1% was 1.1, and BEDROC 0.06. AutoDock failed to retrieve any actives in the top 1% for five targets—ESR1_ago, FEN1, IDH1, OPRK1, and VDR. We can see that IDH1 and OPRK1 targets have completely failed in both algorithms. Success rate was 33%, same as DiffDock, although targets varied. Most notably, the only target was that both algorithms succeeded at was GBA.

**DiffDock-NMDN:** Combining NMDN with DiffDock achieved a median EF1% of 0.67, which is worse than the baseline DiffDock score of 1.17 or DiffDock-GNINA at 0.84. All three algorithms used the same poses, yet DiffDock-NMDN still significantly underperformed compared to the two other scoring approaches. EF1% increased for only three targets: IDH1 (0 → 2.64), ESR1_ant (1.10 → 2.20), and ALDH1 (1.17 → 1.41). Performance worsened on five targets and had no notable impact on the other seven. EF1% fell below 1.0 in eight targets: ADRB2, ESR1_ago, FEN1, MAPK1, OPRK1, PKM2, PPARG, and VDR. These results indicate that DiffDock-NMDN does not outperform DiffDock or DiffDock-GNINA and fails to deliver consistent enrichment gains.

**AutoDock-NMDN**: Rescoring AutoDock–generated poses with NMDN was largely ineffective and, in most cases, detrimental. Median EF1% was 0.37 (worst performing single scoring method). Across targets, EF1% improved for only two targets, FEN1 (0 → 1.63) in and MAPK1 (0.65 → 1.30 in). Both targets were failures in AutoDock-NMDN. Scores declined in seven targets and showed no meaningful change in the remaining six. These results indicate that the NMDN protocol, originally optimized for DiffDock-generated poses, performs considerably better on DiffDock predictions than on AutoDock poses, and that the same protocol produces different results depending on the pose.

**DiffDock-GNINA**: Rescoring DiffDock-generated poses with GNINA resulted Median EF1% of 0.84. It increased EF1% in five targets (versus three DiffDock-NMDN) while four targets showed a decrease. GNINA rescoring achieved EF1% below 1.0 in eight targets, five of them are the same ones that failed with DiffDock-NMDN, which are FEN1, OPRK1, PKM2, PPARG, and VDR. Overall, DiffDock-GNINA did better than DiffDock-NMDN, considering both used same poses for rescoring.

**AutoDock-GNINA:** GNINA rescoring of AutoDock–generated poses emerged as the most effective single-scoring method across both pathways, achieving the highest median EF1% (2.14) and average EF1% (4.07) of all methods. EF1% improved in eight targets, with substantial gains for ESR1_ago (0 → 7.78),

IDH1 (0 → 5.14), and OPRK1 (0 → 12.50). Performance decreased in only two targets—KAT2A (1.55 → 0.52) and MTORC1 (1.03 → 0)—though both showed compensatory EF10% gains and improved BEDROC scores, minimizing negative impact. Importantly, OPRK1, which achieved 0 EF1% across all five other single-scoring methods, was only successfully retrieved with GNINA, highlighting its overall reliability and strong potential to enhance hit identification.

Detailed Per target EF1% performance is shown in Figure 4 across both targets:

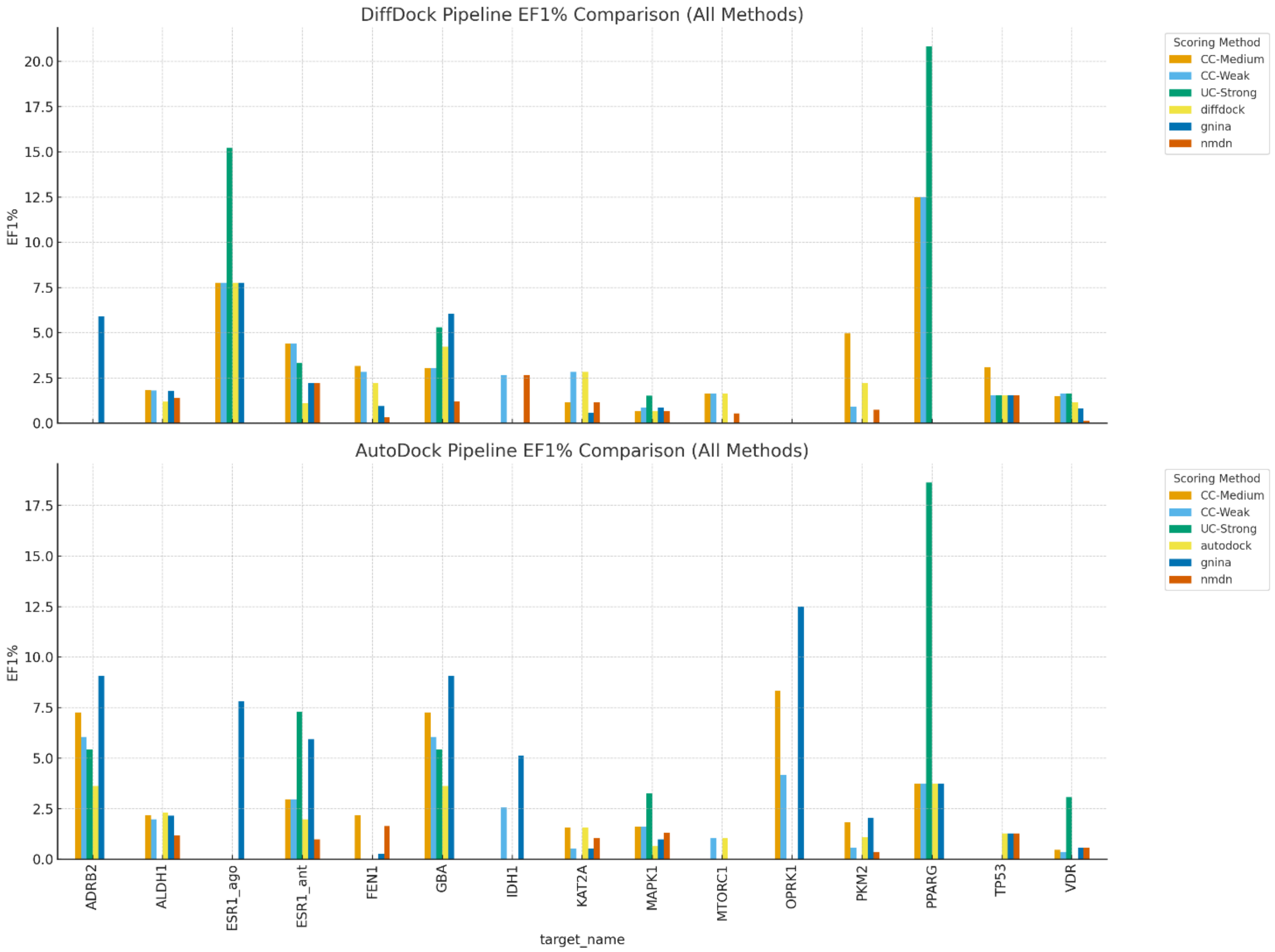

**Figure 4.** *Comparison of EF1% performance across scoring methods for the DiffDock and AutoDock pathways.* This figure shows target-level EF1% values for all scoring methods tested within each pathway. The **top panel** represents DiffDock-generated poses rescored with GNINA, NMDN, and three consensus scoring strategies (CC-Medium, UC-Strong, and CC-Weak). The **bottom panel** represents the same scoring methods applied to AutoDock–generated poses. Overall, both pathways demonstrate substantial variability in performance across targets and scoring methods, with some targets (e.g., **PPARG**, **ESR1_ago**) showing large gains from consensus scoring or rescoring strategies, while others (e.g., **OPRK1**) perform optimally only with GNINA rescoring in the AutoDock pathway. This highlights the complementary strengths of the two pathways and demonstrates the importance of evaluating multiple rescoring strategies to maximize early enrichment performance.

## Performance of consensus methods:

Among the three consensus variants, **CC-Medium** is the most dependable and broadly applicable. It produces nearly identical median EF1% in both pathways (1.84 AutoDock; 1.80 DiffDock) and improves further when rankings are combined globally (1.90 median EF1%). This confirms consensus scoring does indeed improve enrichment. Nevertheless, AutoDock-GNINA remains the strongest *overall* single scoring approach and even compared against consensus, is the per-target winner most often (6/15) in the CC-Medium head-to-head (Table 4).

In table 3 it compares AutoDock-GNINA with other three CC-Medium scores across each pathway (Autodock, Diffodck and GLBOAL)

**Table 4.** Per-target EF1% comparison of AutoDock-GNINA and consensus methods

| Target | AutoDock-GNINA EF1% | AutoDock CC-MEDIUM EF1% | DiffDock CC-Medium EF1% | GLOBAL CC-Medium EF1% |
|---|---|---|---|---|
| ADRB2 | **9.07** | 7.25 | 0 | 5.88 |
| ALDH1 | 2.14 | 2.17 | 1.81 | **2.19** |
| ESR1_ago | **7.82** | 0 | 7.74 | 7.74 |
| ESR1_ant | **5.93** | 2.97 | 4.4 | 2.2 |
| FEN1 | 0.27 | 2.17 | **3.15** | 1.2 |
| GBA | **9.07** | 7.25 | 3.02 | **9.07** |
| IDH1 | **5.1** | 0 | 0 | 0 |
| KAT2A | 0.51 | **1.54** | 1.13 | 0.56 |
| MAPK1 | 0.97 | **1.62** | 0.67 | 0.33 |
| MTORC1 | 0 | **8.33** | 1.62 | 1.625 |
| OPRK1 | **12.5** | 1.84 | 0 | 0 |
| PKM2 | 2.03 | 3.7 | **4.95** | **4.95** |
| PPARG | 3.73 | 3.73 | **12.5** | **12.5** |
| TP53 | 1.28 | 0 | **3.08** | 0 |
| VDR | 0.57 | 0.45 | **1.5** | 0.578 |

| Number of times succeed (score > 1) | 10 (66%) | 11 (73%) | 11 (73%) | 9 (60%) |
|---|---|---|---|---|
| Number of times best scoring method | 6 | 3 | 5 | 4 |

AutoDock-GNINA demonstrates consistent performance across multiple targets compared to consensus approaches.

CC-Weak and UC-Strong delineate the trade-off between coverage and aggressiveness. CC-Weak retains essentially all actives (~100%) while delivering a competitive median EF1% (e.g., ~1.8 in the DiffDock pathway), making it suitable for smaller libraries where maximizing active recovery is a priority. In contrast, UC-Strong can produce spectacular outliers, such as PPARG in Figure 4, which reaches an EF1% of ~20.8 and inflates averages (e.g., DiffDock UC-Strong average EF1% ≈ 3.3), but its median EF1% is 0 and the percentage of actives retained approaches zero, signaling high volatility and limited reliability for prospective campaigns. Taken together, these results suggest a practical strategy: (i) use AutoDock-GNINA as the primary, high-confidence scorer in exploratory campaigns where active chemotypes are unknown; (ii) apply UC-Strong selectively for ultra-large libraries or well-characterized targets where its aggressive filtering can deliver extreme enrichment; and (iii) use CC-Medium as a balanced fallback when GNINA underperforms, offering greater stability and coverage.

**Case study: OPRK1**

The OPRK1 target presented an unusual challenge, as every method in the DiffDock pathway achieved an EF1% of 0, suggesting a potential docking failure. To investigate, we examined the original protein structure and its co-crystallized ligand, using these as references to validate docking accuracy. Visual inspection confirmed that docked ligands were positioned correctly within the expected binding pocket, providing no structural indication for the failure. Refer to figure 5.

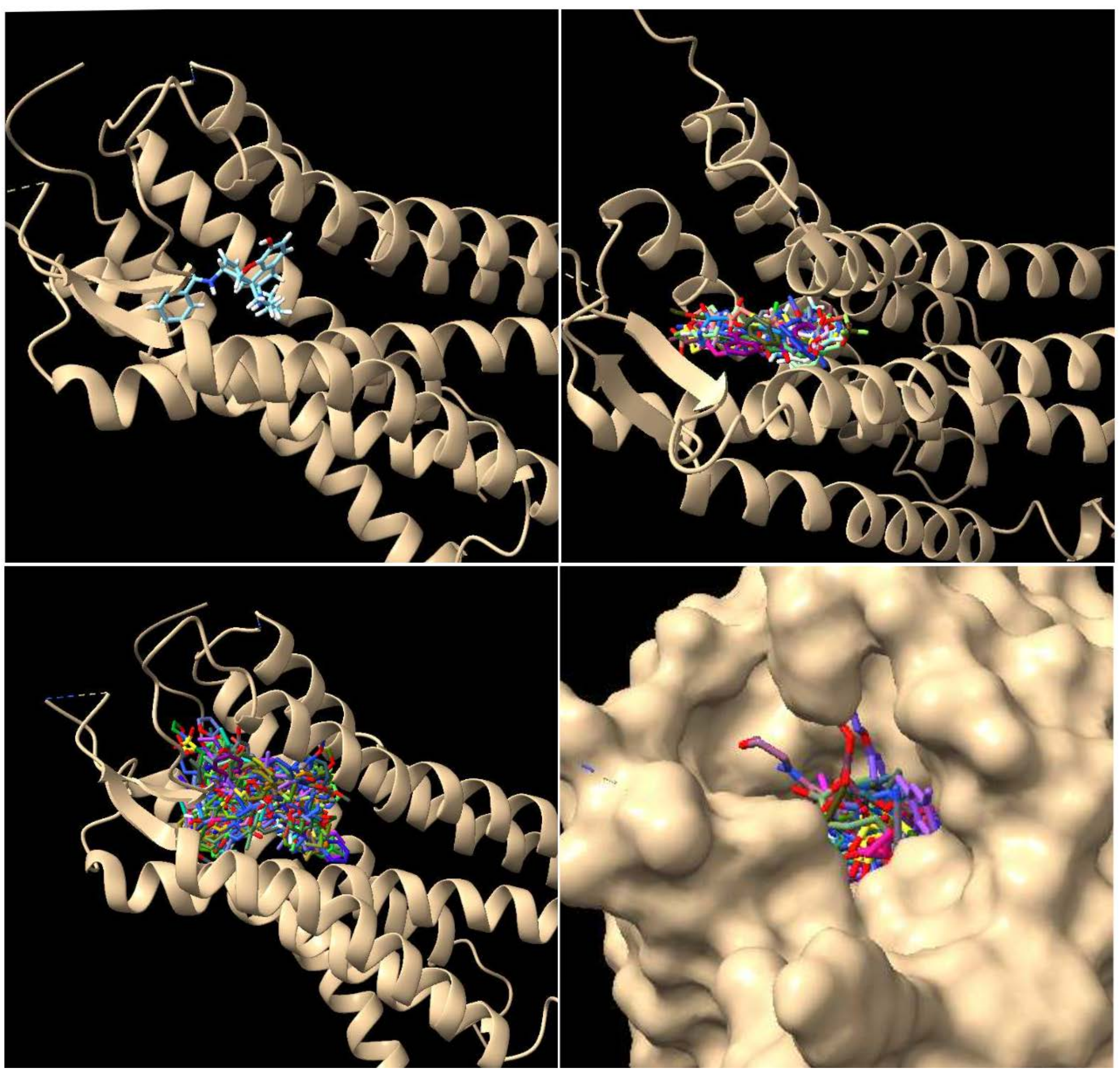

**Figure 5.** OPRK1 protein docking with DiffDock. On the upper left is the original ligand that came with the protein as template for docking. The other 3 pictures illustrate the molecules that have been docked in our pathway.

Despite correct docking poses, visualization of active compound rankings revealed that the 21 actives (out of 197,274 inactives) were concentrated near the bottom of the ranked lists for DiffDock scoring methods. GNINA marginally improved rankings, achieving an EF10% of 0.95, but this translates to needing to screen ~20,000 molecules (top 10%) to retrieve just two actives in a 200,000-compound library, rendering this approach impractical. This case highlights that, for OPRK1, docking performance with the DiffDock pathway was worse than random screening (Figure 6).

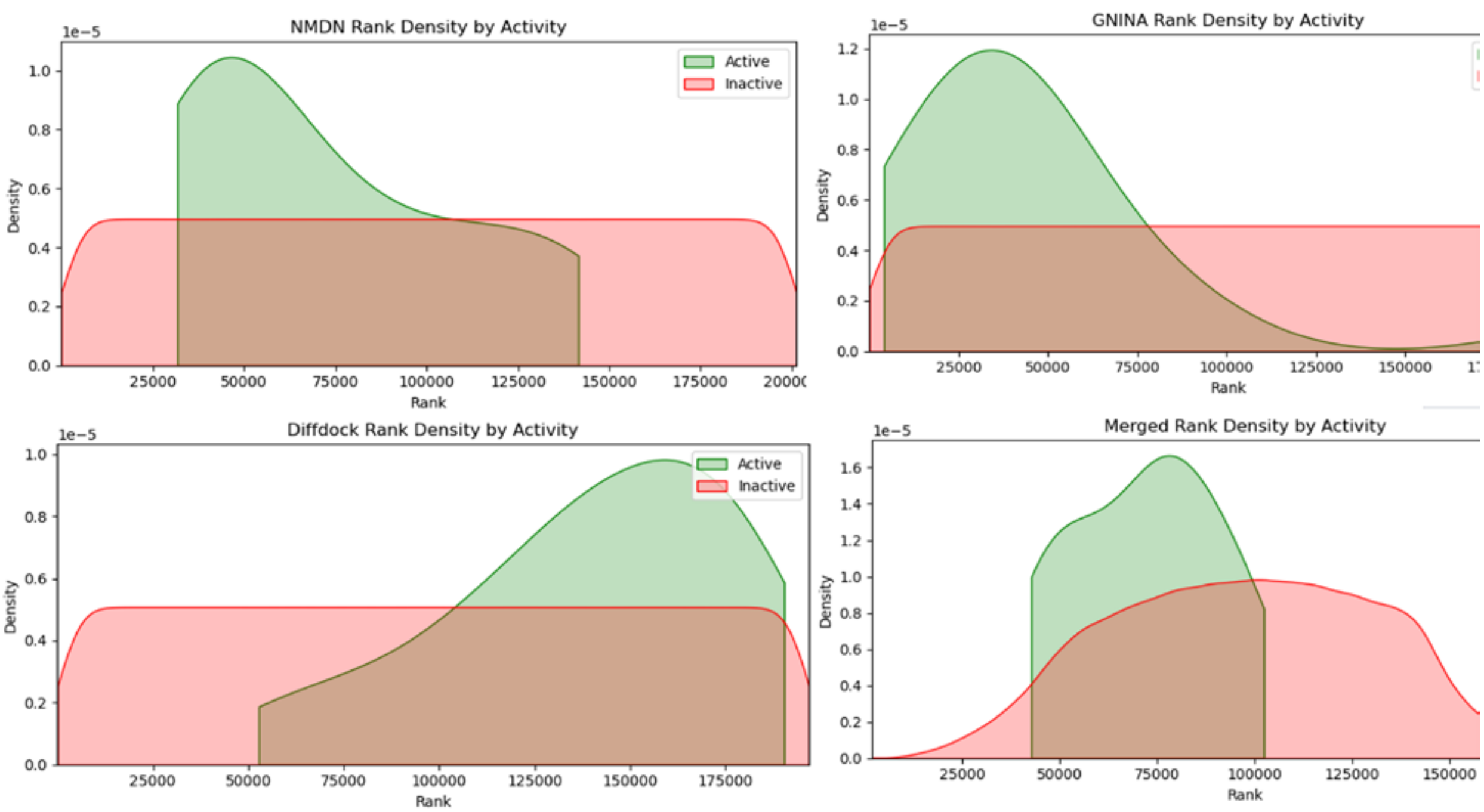

**Figure 6.** Ranking distribution of active compounds across different scoring methods in the DiffDock pathway for OPRK1. GNINA was the only method that managed to shift actives toward the top of the ranking, achieving an EF10% of 0.95.

By contrast, the **AutoDock pathway** demonstrated remarkable reliability for the same target. While baseline AutoDock and AutoDock-NMDN rescoring also yielded EF1% values of 0, **AutoDock-GNINA** dramatically improved performance, achieving an EF1% of 12.5 in a more challenging library of 24 actives and 269,734 inactives. This result underscores the robustness of the AutoDock-GNINA approach, even for large and difficult screening campaigns.

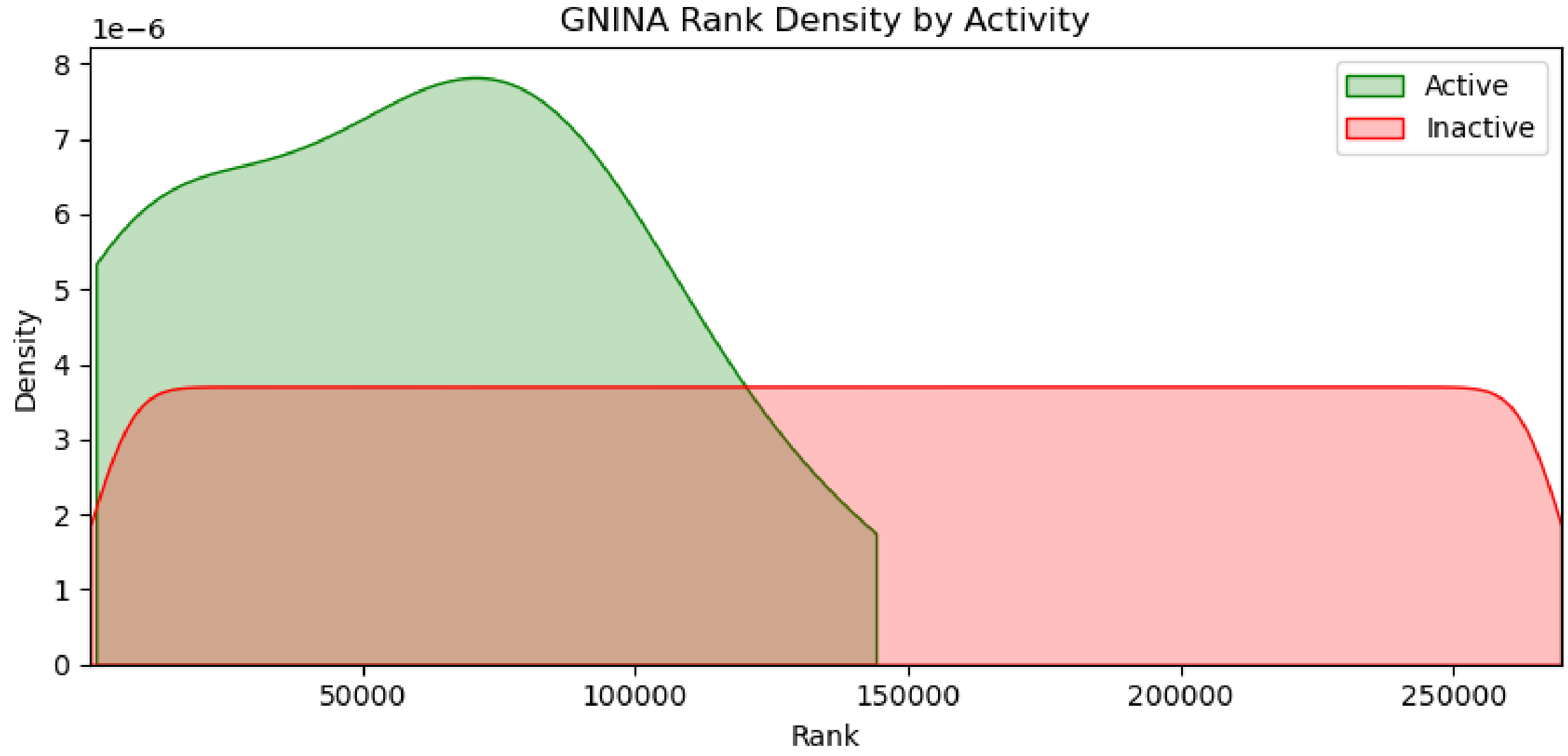

**Figure 7.** Ranking distribution of active compounds by AutoDock-GNINA for OPRK1. Unlike the DiffDock results, it achieved a spectacular EF1% of 12.5 in a library containing 269,734 inactives and 24 actives. This demonstrates GNINA's ability to reliably recover true positives even in challenging large-scale screening scenarios.

**Machine Learning Scoring**:

Learning-to-rank models trained on docking- and rescoring–derived features substantially outperformed the strongest classical single scorer. Using AutoDock-GNINA (EF1% = 2.14) as baseline, the Wide Neural Network (WNN) achieved the best performance with EF1% = 4.494 (+109.8%), more than doubling early enrichment. Detailed results in table 5

Notably, when compared to rank-based averaging consensus strategies, machine learning models still offered clear advantages. Best Consensus method achieved median EF1% values of 1.90 (Global), all well below the WNN's 4.494. This demonstrates that supervised learning can uncover non-linear structure across multiple scoring functions that simple calibrated averaging fails to exploit. Practically, this translates into screening fewer compounds to recover the same number of actives, reinforcing the value of ML re-rankers as an added layer atop established docking rescoring pathway.

Table 5. Performance of Machine Learning Rescoring methods compared to our best traditional docking-scoring method

| Rank | Model | EF@1% | Δ vs Baseline |
|---|---|---|---|
| 1 | Wide Neural Network (512–256–128–1) | 4.49 | +109.8% |
| 2 | Random Forest | 4.10 | +91.6% |
| 3 | XGBoost v2 (200 est., depth 6, LR=0.05) | 3.796 | +77.4% |
| 4 | LightGBM Classification | 3.735 | +74.6% |
| 5 | LambdaMART | 3.279 | +53.2% |
| 6 | Deep Neural Network (256–128–64–1) | 1.792 | –16.3% |
| 7 | AutoDock-GNINA (Baseline) | 2.140 | Ref. |

Wide Neural Network (WNN), more than doubled EF1% compared to the best classical scorer (AutoDock-GNINA, EF1% = 2.14). WNN achieved EF1% = 4.40 (+105.7%), with Deep MLP and XGBoost variants also showing substantial gains. These results confirm that ML-based rescoring capture richer patterns than rank-based consensus methods (median EF1% ≈ 1.8–1.9), offering significant efficiency gains in prioritizing compounds for experimental testing.

## Classical Accuracy Metrics:

To make our findings more accessible to readers outside the docking and cheminformatics domain, we also reported performance using conventional machine learning metrics such as **accuracy, precision, recall,** **and F1-score**. This helps illustrate, in simple terms, how effective (or ineffective) docking-based methods are at identifying true actives within massive chemical libraries.

If we assumed the top 1% of the ranked molecules are actives, and all the rest are not, AutoDock-GNINA rescoring will achieve a median accuracy of 98.1%, which at first glance appears strong. However, this figure is misleading because of the overwhelming dominance of inactives in the dataset. Precision, on the other hand was only 1.85%, meaning that fewer than 2 of every 100 compounds predicted as actives were truly active. Recall was 2.02%, showing that only about 2 in every 100 true actives were successfully identified. The corresponding F1-score was 1.9%. A more balanced accuracy measure is calculated using equation 3.

$$Balanced\ Accruacy = \frac{Sensetivity + Specifitiy}{2}$$

Equation 3

The result is 50.53%, which is slightly better than random guessing.

At its optimal decision threshold (0.1), the WNN achieved a median accuracy of 93.9%. Precision was 8.2%, indicating that ~8 of every 100 predicted actives were true hits—over four times higher than GNINA's 1.85%. Recall reached 15.0%, showing that the WNN successfully retrieved 15 of every 100 true actives, compared to only 2 for GNINA. The corresponding F1-score was 10.6%, with a Balanced Accuracy of 55.4%—notably higher than GNINA's 50.5% and well above random guessing. The Matthews Correlation Coefficient (MCC), a stricter balanced metric, was 0.081, again confirming improvement over the near-random baseline.

## Discussion

This study was designed to systematically evaluate some of the new techniques such as DiffDock and DiffDock-NMDN and compare it with some of the classical, physics-based methods such as AutoDock. Our findings reveal a nuanced landscape where newer, more complex models do not axiomatically outperform well-established methods.

The combination of AutoDock and GNINA (AutoDock-GNINA) not only achieved the highest median EF1% (2.03) of any individual method but also showed much faster inference. With AutoDock it processes about 2 ligands per second on an RTX 3090. On the other hand, DiffDock takes about 1-2 seconds to process one ligand on an A100, which by itself has at least double performance of an RTX 3090. [15, 16] That means that it probably takes 4 to 8 times computer power to process ligand with DiffDock compared to AutoDock, just to deliver inferior precision eventually.

The AutoDock baseline itself, however, performed poorly, often failing to enrich actives better than random chance. This confirms the well-documented limitations of traditional empirical scoring functions. In a study by Yu et al. (2023), AutoDock's performance achieved an $EF_1$% of zero in five of the eight targets. [17] However, its strength lies in its fast and efficient conformational sampling algorithm. It appears that by providing a diverse and geometrically plausible set of ligand poses. GNINA, with its 3D convolutional neural network architecture, excels at re-evaluating these poses, effectively identifying the subtle structural and chemical features indicative of a true binding event. This is confirmed by the significant difference in performance of GNINA with DiffDock which was 0.84 Median EF1% compared with its performance in the AutoDock pathway that was almost double the 2.14 EF1%.

AutoDock and GNINA's combination was also not perfect; it completely failed against certain targets such as FEN1, where both EF1% and EF10% were below 1, and the ROC-AUC was only 0.509, indicating near-random ranking performance. This aligns with Zhang et al. (2023), who reported that Glide SP achieved EF1% = 0 on LIT-PCBA targets including FEN1 [18] On the other hand, Xia et al., 2025 demonstrated that DiffDock-NMDN-generated poses allowed multiple scoring functions to achieve more meaningful enrichment for FEN1 (e.g., AD4 = 1.08, Vina = 2.17, Vinardo = 1.90, NMDN = 2.44, and combined $pK_{(d)}$+NMDN = 3.79). [13] This is also not far from our results, were NMDN rescoring on the same poses generated by AutoDock achieved EF1% of 1.63 compared to the 0.27 of GNINA, indicating that may be inherited supremacy related to NMDN scoring in that particular target.

The same NMDN success with AutoDock poses on FEN1 could not be replicated with DiffDock. Again, this may be related to the poses generated by DiffDock. One example is molecule 17434066 (figure 8) with smile "CCOc1cc(\C=C\2/N=C(SCC=C)SC2=O)cc(c1OC(=O)C)[N+](=O)[O-]". On the right is the pose that was generated by AutoDock and NMDN gave it a pKd-Score 46.62 so it ranked number 7 in the list. On left, is the same molecule pose that was generated by DiffDock it was given a pKd-Score of just 3.19. So it ranked 7039 out 18135. This illustrates that pose quality is a primary determinant of machine-learning rescoring outcomes. While scoring 20–40 poses per ligand (as proposed by Xia et al. [13]) could mitigate single-pose errors, it would increase compute by ~20–40× relative to our setup—posing a practical limitation for real-world, large-scale screening.

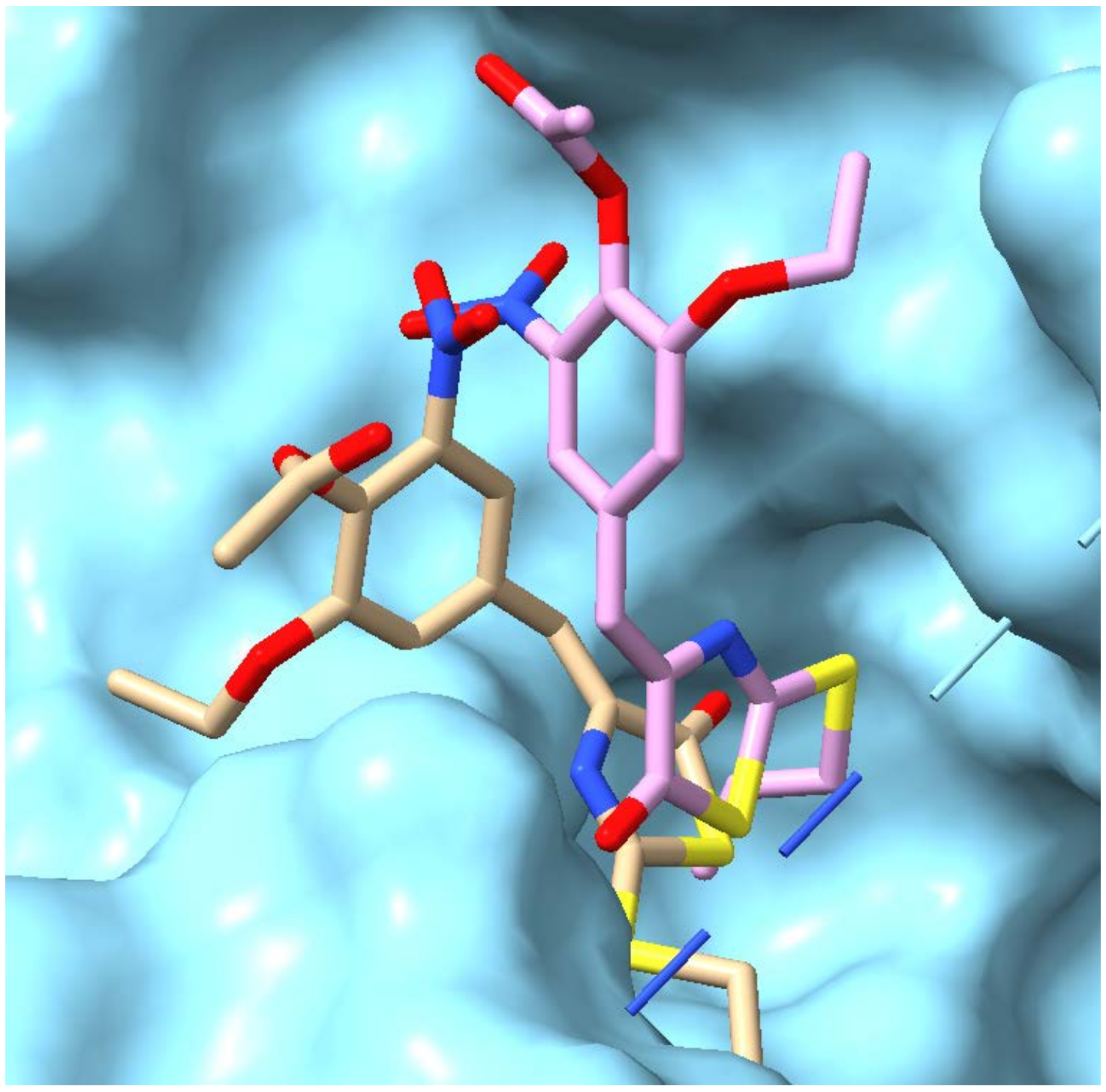

**Figure 8**. the active molecule "17434066". On the right is the pose that was generated by AutoDock and successfully ranked by NMDN, and on the left is the pose that was generated by DiffDock and falsely ranked.

**The reappearing concept of consensus docking**

Other researchers have attempted various implementations of Conesus docking. The principle is that while any single scoring function may perform poorly on a specific target or chemical series, its errors are often idiosyncratic. By intelligently aggregating the scores or rankings from multiple, diverse docking methods, these individual errors can be averaged out, leading to a final consensus score that is more accurate, robust, and generalizable than any of its constituent parts. [2] This approach seeks to improve the signal-to-noise ratio in virtual screening hit lists, thereby enhancing the overall predictive power.

**Morris (2022)** introduced MILCDock, a method that employs a Multi-Layer Perceptron (MLP) to generate a consensus prediction from 52 features derived from five different docking tools. The model was trained on the labeled active and inactive compounds from the DUD-E and LIT-PCBA datasets to classify potential ligands. [2]

MILCDock achieved an average EF1% of 4.37—comparable to the median EF1% of 4.49 in our study. However, training models directly on benchmarking datasets poses major limitations, chiefly poor generalizability and reliance on artificial decoys. Models trained on specific targets often overfit the specific chemical space of the benchmarks, failing to generalize to new protein targets. Consequently, this supervised ML approach has to be only considered for targets with already existing data or at least similar targets.

Another study by Lacour et al. introduced DockM8, an open-source consensus virtual screening platform that emphasizes target-specific customization over universal prediction [19]. Rather than training a single model, DockM8 integrates multiple docking engines, sixteen scoring functions, and ten consensus strategies in a modular workflow that researchers can tune per target. This adaptability led to state-of-the-art results on the challenging LIT-PCBA dataset, where DockM8-max achieved a median EF1% of 7.83 and AUROC of 0.623, outperforming eighteen competing protocols. The authors highlight that no single method dominates across all targets. They propose using DeepCoy-generated decoys to benchmark workflows when a few known actives exist, allowing researchers to identify an optimal setup before large-scale screening.

Despite these strengths, DockM8's core limitation mirrors its advantage—its high degree of customizability demands prior knowledge and extensive computation. The framework performs best when benchmark actives and inactives are available, but many early-stage discovery efforts lack such data, making optimization difficult. Even with decoy generation, the approach assumes access to at least some validated binders. Moreover, reaching per-target optima required exploring roughly $6.2 \times 10^5$ workflow combinations, imposing heavy computational costs. In practice, achieving DockM8's full potential necessitates substantial compute resources, expert oversight and availability of suitable dataset, which is often challenging.

As of early 2026, several new AI-driven drug discovery frameworks have been introduced that aim to address some of the long-standing limitations of docking-based virtual screening. One of the most recent examples is the Isomorphic Labs Drug Design Engine (IsoDDE), a unified deep-learning system that integrates biomolecular structure prediction, protein–ligand binding pose generation, and binding affinity prediction, within a single framework. [20] According to their initial technical reports, IsoDDE demonstrates substantial improvements over previous systems such as AlphaFold 3 in predicting protein–ligand structures and binding affinities. Historically, many newly introduced AI-based docking and scoring tools have reported strong performance on curated benchmarks but failed to reproduce those gains in more realistic screening environments. Rigorous independent evaluation on experimentally derived datasets such as LIT-PCBA and in prospective drug discovery campaigns will therefore be essential before the reliability and generalizability of these new systems can be established.

**The Limitations of Docking in Drug Discovery**

Even the best-performing components in our benchmark expose inherent limits of docking for prospective hit finding. First, there is a persistent gap between benchmark gains and real-world efficacy. Methods that look strong on decoy-based suites (e.g., CASF) can contract sharply on experimentally curated sets like LIT-PCBA. We see the same pattern here: our strongest classical single scorer, AutoDock-GNINA, delivers only median EF1% 2.14 and ROC-AUC values hovering near 0.6, i.e., modest discrimination. In classical accuracy/precision terminology, labeling the top 1% as “predicted actives”, GNINA’s median precision is only 1.85% recall 2.02, and balanced accuracy of 50.5% which is barely above random prediction (50%).

Second, performance is highly target-dependent. Even our most successful and robust scoring methods “consensus scoring” had EF1% less than 1 in 4 out of 15 targets, while AutoDock-GNINA, had EF1% of less than 1 in 5 different targets, which is even worse than randomly screening. This indicates that there is an inherit problem in these docking/scoring methods that make them unable to generalize beyond certain protein targets.

Third, the question of which pose should be used, in our method we used the ones which had highest confidence score with DiffDock and the ones that achieved highest binding affinity with AutoDock-GNINA. What if different strategy for pose selection is used, such as selecting the most common pose, or the ones that are closest to the true binder. Will different pose strategy help scoring methods achieve better performance? Maybe. But it also adds to the complexity of docking, and the uncertainty of any approach.

Fourth, consensus and ML help—but do not fully solve the problem. Our best Calibrated rank consensus (CC-Medium) raised robustness in AutoDock slightly. Succeeded in one extra target versus AutoDock-GINA (11 vs 10) but because of its complexity and computational demand it may not be justifiable for usage. Supervised models went further: our Wide Neural Network (WNN) achieved EF@1% = 4.49 (≈+110% vs GNINA) and improved classical metrics at an F1-optimal threshold (precision 8.2%, recall 15.0%, balanced accuracy 55.4%). However, training ML models risk overfitting and limited generalization to novel targets/chemotypes.

Finally, new evidence from systems pharmacology suggests deeper limitations in structure-based approaches. Abo-Dahab *et al.* (2026) showed that pharmacology knowledge graphs can effectively predict drug–protein interactions and enable drug repurposing using only network topology and target-centric features (e.g., protein embeddings), with explicit chemical structure representations (such as Morgan fingerprints or graph attention encoders) proving not only redundant but often detrimental to performance. This implies that relying heavily on atomic-level structural information — the very foundation of molecular docking and pose scoring — may carry inherent constraints that cannot be fully overcome by better sampling or rescoring alone. [21]

These limitations indicate that docking while still have its advantages, cannot be solely relied on for drug discovery.

## Limitations of Our Study

First, our design emphasized single-pose evaluation for each ligand to keep pathways comparable and computationally tractable at LIT-PCBA scale. Many methods—especially diffusion/search-based workflows—benefit from multi-pose sampling plus ensemble rescoring. By not exploiting pose ensembles, we likely underestimate ceiling performance for these methods.

Second, we applied 5% inactives subsampling for several targets (while keeping all actives) to fit compute budgets. Subsampling preserves rank order in expectation but can shift EF statistics and tail behavior, particularly for EF1% where a few actives dominate variance. Reported medians and averages should therefore be read as conservative approximations for those targets.

Third, our consensus schemes (cutoffs and weights) were simple and fixed across targets. This choice was to measure robustness but may fall short of target-specific optima. More aggressive per-target calibration, threshold optimization, and cost-sensitive tuning could yield additional gains at the expense of complexity.

Finally, Our findings are specific to the LIT-PCBA dataset, which, while far more realistic than decoy-based benchmarks, is still just one compilation of targets and assays. Caution is warranted in generalizing conclusions to other contexts. Different protein families, different ligand libraries (e.g., proprietary corporate collections), or different preparation protocols (ligand protonation, protein pocket definition, etc.) could yield different outcomes. We focused on LIT-PCBA because it offers a large-scale, public, and relatively unbiased testbed. However, the absolute performance numbers (and even the relative ranking of methods) might shift in other scenarios. For instance, in an easier benchmark with analog biases, DiffDock or ML scorers might shine more; in a completely novel target with no similar training data, they might fare worse. Thus, while we believe the trends observed (e.g., value of GNINA rescoring, pitfalls of DiffDock on novel data) are informative, they are not universal truths for all virtual screening efforts.

## Conclusions

Docking-based virtual screening provides useful enrichment but limited absolute predictive power on realistic datasets such as LIT-PCBA. Among the evaluated approaches, AutoDock-GNINA was the most reliable single method, while calibrated consensus improved robustness without surpassing peak enrichment. Supervised machine-learning re-ranking delivered the largest gains, more than doubling early enrichment relative to classical scoring but its generalizability not certain. Overall, effective workflows should only use methods that can be validated against that specific target or closely related systems.

## Declarations

### Availability of data and materials

The LIT-PCBA dataset used in this study is publicly available from the original publication.

The datasets generated and analyzed during the current study (including docking results, rescoring outputs, and machine learning features) are available from the corresponding author upon reasonable request.

### Competing interests

The authors declare that they have no competing interests.

### Funding

This work was supported by internal funding from the Zhao Lab (Dr. Liang Zhao) at the University of California, San Francisco.

Additional support was provided through computational resources from the Zhao Lab.

### Authors' contributions

Y.A. conceived and designed the study, performed all computational experiments, analyzed the data, and wrote the manuscript.

L.Z. supervised the study and provided strategic guidance.

X.X. and J.C. contributed to manuscript review and editing.

All authors read and approved the final manuscript.

### Acknowledgements

This work was conducted as part of a capstone project for Master of Science degree in Artificial Intelligence and Computational Drug Discovery and Development (AICD3) at the University of California, San Francisco.

The authors acknowledge the Zhao Lab for providing computational resources and support.

References:

1. Sunseri J, Koes DR. Virtual screening with Gnina 1.0. Molecules. 2021;26(23):7369. doi:10.3390/molecules26237369.

2. Morris CJ. MILCDock: Machine Learning-Enhanced Consensus Docking for Virtual Screening in Drug Discovery [senior thesis]. Provo (UT): Brigham Young University; 2022. Available from: https://physics.byu.edu/docs/thesis/1542
3. da Silva MMP, Guedes IA, Custódio FL, Krempser E, Dardenne LE. Deep learning strategies for enhanced molecular docking and virtual screening. ChemRxiv [Preprint]. 2023 Nov 7. doi:10.26434/chemrxiv-2023-zfv87-v2.
4. Mysinger MM, Carchia M, Irwin JJ, Shoichet BK. Directory of useful decoys, enhanced (DUD-E): better ligands and decoys for better benchmarking. J Med Chem. 2012;55(14):6582-6594. doi:10.1021/jm300687e.
5. Tran-Nguyen VK, Jacquemard C, Rognan D. LIT-PCBA: an unbiased data set for machine learning and virtual screening. J Chem Inf Model. 2020;60(9):4263-4273. doi:10.1021/acs.jcim.0c00155.
6. Huang A, Knight IS, Naprienko S. Data leakage and redundancy in the LIT-PCBA benchmark. arXiv [Preprint]. 2025 Jul 29; arXiv:2507.21404. Available from: https://arxiv.org/abs/2507.21404
7. Truchon JF, Bayly CI. Evaluating virtual screening methods: good and bad metrics for the “early recognition” problem. J Chem Inf Model. 2007;47(2):488-508. doi:10.1021/ci600426e.
8. Solis-Vasquez L, Tillack AF, Santos-Martins D, Koch A, LeGrand S, Forli S. Benchmarking the performance of irregular computations in AutoDock-GPU molecular docking. Parallel Comput. 2022;109:102861. doi:10.1016/j.parco.2021.102861.
9. Darme P, Dauchez M, Renard A, Voutquenne-Nazabadioko L, Aubert D, Escotte-Binet S, et al. AMIDE v2: high-throughput screening based on AutoDock-GPU and improved workflow leading to better performance and reliability. Int J Mol Sci. 2021;22(14):7489. doi:10.3390/ijms22147489.
10. Santos-Martins D, Solis-Vasquez L, Tillack AF, Sanner MF, Koch A, Forli S. Accelerating AutoDock4 with GPUs and gradient-based local search. J Chem Theory Comput. 2021;17(2):1060-1073. doi:10.1021/acs.jctc.0c01006.
11. Corso G, Stärk H, Jing B, Barzilay R, Jaakkola T. DiffDock: diffusion steps, twists, and turns for molecular docking. arXiv [Preprint]. 2022 Oct 4; arXiv:2210.01776. Available from: https://arxiv.org/abs/2210.01776
12. Jain AN, Cleves AE, Walters WP. Deep-learning based docking methods: fair comparisons to conventional docking workflows. arXiv [Preprint]. 2024 Dec 3; arXiv:2412.02889. Available from: https://arxiv.org/abs/2412.02889.
13. Xia S, Gu Y, Zhang Y. Normalized protein-ligand distance likelihood score for end-to-end blind docking and virtual screening. J Chem Inf Model. 2025;65(3):1101-1114. doi:10.1021/acs.jcim.4c01014.
14. Nahm FS. Receiver operating characteristic curve: overview and practical use for clinicians. Korean J Anesthesiol. 2022;75(1):25-36. doi:10.4097/kja.21209.
15. Tang S, Chen R, Lin M, Lin Q, Zhu Y, Ding J, et al. Accelerating AutoDock Vina with GPUs. Molecules. 2022;27(9):3041. doi:10.3390/molecules27093041.
16. Bizon-Tech. NVIDIA RTX 3090 vs NVIDIA A100 40 GB (PCIe): GPU benchmarks [Internet]. [cited 2025 Aug 27]. Available from: https://bizon-tech.com/gpu-benchmarks/NVIDIA-RTX-3090-vs-NVIDIA-A100-40-GB-(PCIe)/579vs592

17. Yu Y, Cai C, Wang J, Bo Z, Zhu Z, Zheng H. Uni-Dock: GPU-accelerated docking enables ultralarge virtual screening. J Chem Theory Comput. 2023;19(11):3336-3345. doi:10.1021/acs.jctc.2c01145.
18. Zhang X, Shen C, Jiang D, et al. TB-IECS: an accurate machine learning-based scoring function for virtual screening. J Cheminform. 2023;15:63. doi:10.1186/s13321-023-00731-x.
19. Lacour A, Ibrahim H, Volkamer A, Hirsch AKH. DockM8: an all-in-one open-source platform for consensus virtual screening in drug design. ChemRxiv [Preprint]. 2024 Jul. doi:10.26434/chemrxiv-2024-17k46.
20. Isomorphic Labs Team. Accurate predictions of novel biomolecular interactions with IsoDDE. Zenodo [Report]. 2026 Feb 10. doi:10.5281/zenodo.18606681.
21. Abo-Dahab, Y., Hernandez, R. & Duran, I.C.A. Pharmacology knowledge graphs enable drug repurposing without chemical structure information. Discov Artif Intell (2026). https://doi.org/10.1007/s44163-026-01303-2